\documentclass[10pt,twocolumn,letterpaper]{article}

\usepackage{cvpr}
\usepackage{times}
\usepackage{epsfig}
\usepackage{graphicx}
\usepackage{amsmath}
\usepackage{amssymb}

\usepackage{xcolor}
\usepackage{subcaption}

\usepackage[breaklinks=true,bookmarks=false]{hyperref}

\cvprfinalcopy 

\def\cvprPaperID{12} 

\ifcvprfinal\fi
\begin{document}

\title{Filmy Cloud Removal on Satellite Imagery with\\ Multispectral Conditional Generative Adversarial Nets}

\author{Kenji Enomoto$^{1*}$ \ Ken Sakurada$^{1}$ \ Weimin Wang$^{1}$ \ Hiroshi Fukui$^{2}$ \ Masashi Matsuoka$^{3}$\\ Ryosuke Nakamura$^{4}$ \ Nobuo Kawaguchi$^{1}$\\
$^1$Nagoya University \ $^2$Chubu University \ $^3$Tokyo Institute of Technology\\ $^4$Advanced Industrial Science and Technology\\
{\tt\small \{enoken, weimin\}@ucl.nuee.nagoya-u.ac.jp, \{sakurada, kawaguti\}@nagoya-u.jp}\\ {\tt\small fhiro@vision.cs.chubu.ac.jp, matsuoka.m.ab@m.titech.ac.jp, r.nakamura@aist.go.jp}
}


\maketitle

\begin{abstract}
In this paper, we propose a method for cloud removal from visible light RGB satellite images by extending the conditional Generative Adversarial Networks (cGANs) from RGB images to multispectral images. Satellite images have been widely utilized for various purposes, such as natural environment monitoring (pollution, forest or rivers), transportation improvement and prompt emergency response to disasters. However, the obscurity caused by clouds makes it unstable to monitor the situation on the ground with the visible light camera. Images captured by a longer wavelength are introduced to reduce the effects of clouds. Synthetic Aperture Radar (SAR) is such an example that improves visibility even the clouds exist. On the other hand, the spatial resolution decreases as the wavelength increases. Furthermore, the images captured by long wavelengths differs considerably from those captured by visible light in terms of their appearance. Therefore, we propose a network that can remove clouds and generate visible light images from the multispectral images taken as inputs. This is achieved by extending the input channels of cGANs to be compatible with multispectral images. The networks are trained to output images that are close to the ground truth using the images synthesized with clouds over the ground truth as inputs. In the available dataset, the proportion of images of the forest or the sea is very high, which will introduce bias in the training dataset if uniformly sampled from the original dataset. Thus, we utilize the t-Distributed Stochastic Neighbor Embedding (t-SNE) to improve the problem of bias in the training dataset. Finally, we confirm the feasibility of the proposed network on the dataset of four bands images, which include three visible light bands and one near-infrared (NIR) band.
\end{abstract}

\section{Introduction}

\begin{figure}[!t]
\begin{center}
\hspace*{0mm}\includegraphics[width=80mm,bb=0 0 459 299]{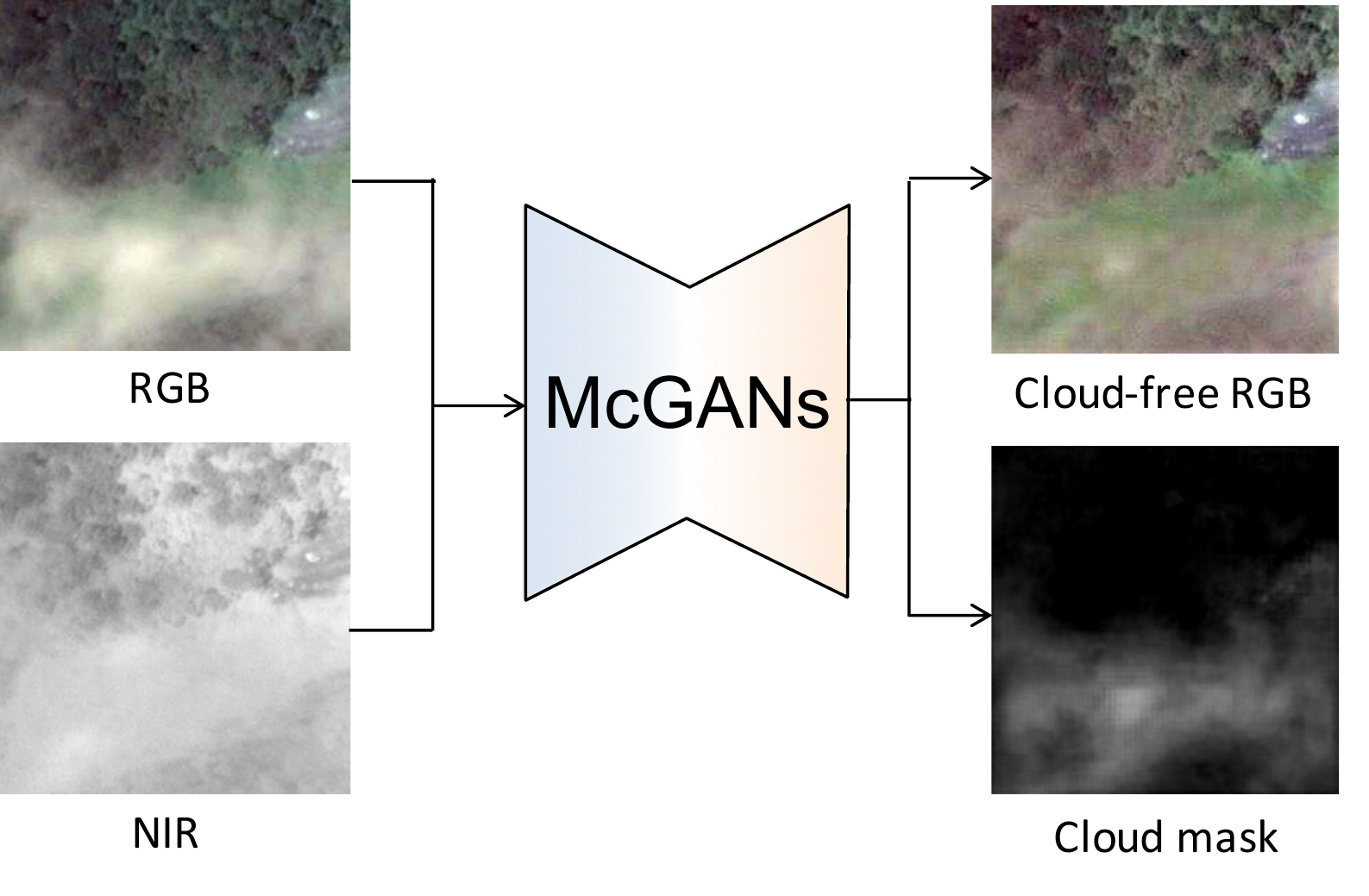}
\caption{McGANs for cloud removal}
\label{fig:abst}
\end{center}
\end{figure}

Satellite images have been widely utilized in various of fields such as remote sensing, computer vision, environmental science and meteorology.
With the help of satellite images, we can observe the situation on the ground for natural environment monitoring (pollution, forest or rivers), transportation improvement and prompt emergency response to disasters.
There are many research area dealing with satellite images, e.g., object recognition from the satellite images, change detection for ground usage or disaster situation analysis.

However, the obscurity caused by the cloud makes it unstable to monitor the situation on the ground with a visible light camera. To be unaffected by the cloud, images captured by longer wavelengths are introduced. Synthetic Aperture Radar (SAR) \cite{furuta_sar_color} is such an example, which improves visibility even in the presence of clouds. On the other hand, the spatial resolution decreases as the wavelength increases. Furthermore, the image captured by a long wavelength differs considerably in appearance from the one captured by visible light. This affects the visibility for observation.

In this paper, we propose Multispectral conditional Generative Adversarial Networks (McGANs) based on conditional Generative Adversarial Networks（cGANs), for cloud removal from visible light RGB satellite images with multispectral images as inputs. See Fig.\ref{fig:abst} for illustration. Compared with cGANs, the input channels of McGANs are expended for multispectral images. For the input of RGB images obscured by clouds and the registered NIR images, McGANs is trained to output the RGB images that are close to the ground truth. However, it is impractical to capture the cloud-free and the cloud obscured images of the completely same scene at the same time. Hence, we synthesize images with the simulated clouds over the ground truth RGB images to generate the training data. Furthermore, the prediction accuracy is expected to be improved by training the networks to detect the region of cloud simultaneously. Both the synthesized and the ground truth RGB images are color corrected to eliminate the affection of color tone caused by variety of imaging conditions such as weather, lighting and the processing method of the image sensor. 

In the available dataset, the ratio of images of the forest or the sea is very high, which will introduce bias in the training dataset if uniformly sampled from the original dataset. Thus, we utilize the t-Distributed Stochastic Neighbor Embedding (t-SNE) \cite{maaten2008visualizing} to reduce the bias problem of the training dataset. Finally, we confirm the feasibility of the proposed networks on the dataset of four bands images, which includes three visible light bands and one near-infrared (NIR) band.

\section{Related Work}
In the field of remote sensing, microwave is usually utilized since it is unaffected by the cloud cover. Synthetic Aperture Radar (SAR) is mounted on airplanes and satellites to overcome the shortage of low spatial resolution of the microwave. Nonetheless, the resolution of SAR images is still much lower than that of the images captured by visible light. Besides, it is difficult to understand the SAR images directly. To improve the visibility of SAR images, there also exists the work about coloring these SAR images \cite{furuta_sar_color}. 

In the field of computer vision, many dehazing methods have been proposed for  RGB images only \cite{he2011single,berman2016non} or for both RGB and NIR  images \cite{schaul2009color,feng2013near,shibata2015unified}. The pre-knowledge or assumption about the color information of the hazed imaged is necessary in the former method. In the latter, NIR images, which possess higher penetrability through fog than the visible light, are used as the guide to dehaze the RGB images.

Generative Adversarial Networks~(GANs)~\cite{goodfellow2014generative} is the most relevant to our work. GANs is consisted of two types of networks, Generator and Discriminator. Generator is trained to generate images that cannot be discriminated by Discriminator with the ground truth, while Discriminator is trained to discriminate between the generated images by the Generator and the ground truth. The conditional version of GANs was also proposed in \cite{mirza2014conditional}. However, learning by GANs is unstable. To increase the stability, convolutional networks and  Batch Normalization  are introduced to Deep Convolutional Generative Adversarial Networks (DCGANs) \cite{radford2015unsupervised} is proposed.

Research about image generation based on cGANs and DCGANS has been widely applied for image restoration or the removal of certain objects such as rain and snow \cite{pathak2016context,zhang2017image}. In particular, the method in \cite{isola2016image} can generate general and high-quality images by combing Generator of U-Net \cite{ronneberger2015u} and  Discriminator of PatchGAN~\cite{li2016precomputed}. The Generator of U-Net spreads the missing spatial features in the convolution layers of Encoder to each layer of Decoder by introducing the skip connection between layers of Encoder and Decoder. PatchGAN is able to model the high frequencies for sharp details by training the Discriminator on the image patches. Generally, these cGANs-based methods predict the obscured regions of the image with the surrounding unobscured information only from the input RGB images. 

Based on the aforementioned research, we propose the cloud removal networks by taking the advantage of the color information from visible light images and the high penetrability from images captured by longer wave. The proposed networks predict the obscured region from not only the RGB images but also images captured by longer wavelengths that can partly or completely penetrate the cloud. Our final purpose is to implement the networks that can merge SAR images  captured by the cloud-penetrating microwave. As the first step, we construct and evaluate the networks for cloud removal with the visible light RGB images and the near-infrared spectrum NIR images in this work (the region of NIR  wavelength is the closest to visible light). 

\begin{figure*}[!t]
\begin{center}
$\begin{array}{p{40mm}p{40mm}p{40mm}p{40mm}}
\hspace*{0mm}\includegraphics[width=40mm,bb=0 0 256 256]{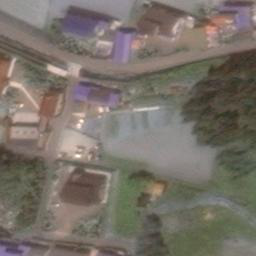}\subcaption{}\label{fig:synthesis:rgb}
&\hspace*{0mm}\includegraphics[width=40mm,bb=0 0 256 256]{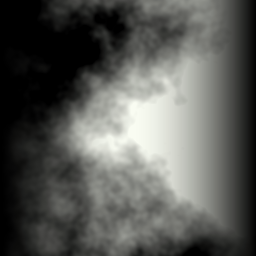}\subcaption{}\label{fig:synthesis:cloud}
 &\hspace*{0mm}\includegraphics[width=40mm,bb=0 0 256 256]{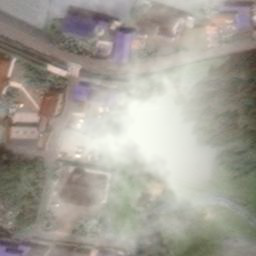}\subcaption{}\label{fig:synthesis:syn}\
 &\hspace*{0mm}\includegraphics[width=40mm,bb=0 0 256 256]{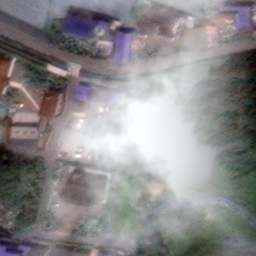}\subcaption{}\label{fig:synthesis:cc}
\end{array}$

\caption{Synthesis of cloud obscured images. \protect\subref{fig:synthesis:rgb}: Original RGB image. \protect\subref{fig:synthesis:cloud}: Simulated cloud using Perlin noise. \protect\subref{fig:synthesis:syn}: Merged image with the cloud by alpha blending. \protect\subref{fig:synthesis:cc}: Final result after color correction}
\label{fig:synthesis}

\end{center}
\end{figure*}

\begin{figure}[!t]
\begin{center}

$\begin{array}{p{40mm}p{40mm}}
\hspace*{0mm}\includegraphics[width=40mm,bb=0 0 256 256]{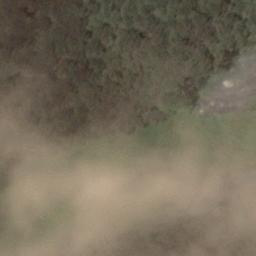}\subcaption{}\label{fig:synthesis_colorcorrect:before}
&\hspace*{0mm}\includegraphics[width=40mm,bb=0 0 256 256]{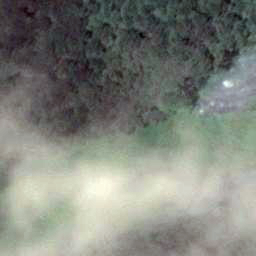}
\subcaption{}\label{fig:synthesis_colorcorrect:after}
\end{array}$

\caption{Example of color correction for real RGB image with cloud. \protect\subref{fig:synthesis_colorcorrect:before}: Original RGB image obscured by the cloud. \protect\subref{fig:synthesis_colorcorrect:after}: Color corrected result.}
\label{fig:synthesis_colorcorrect}

\end{center}
\end{figure}

\begin{figure*}[!t]
\begin{center}

$\begin{array}{p{80mm}p{80mm}}
\hspace*{0mm}\includegraphics[width=80mm,bb=0 0 2000 2000]{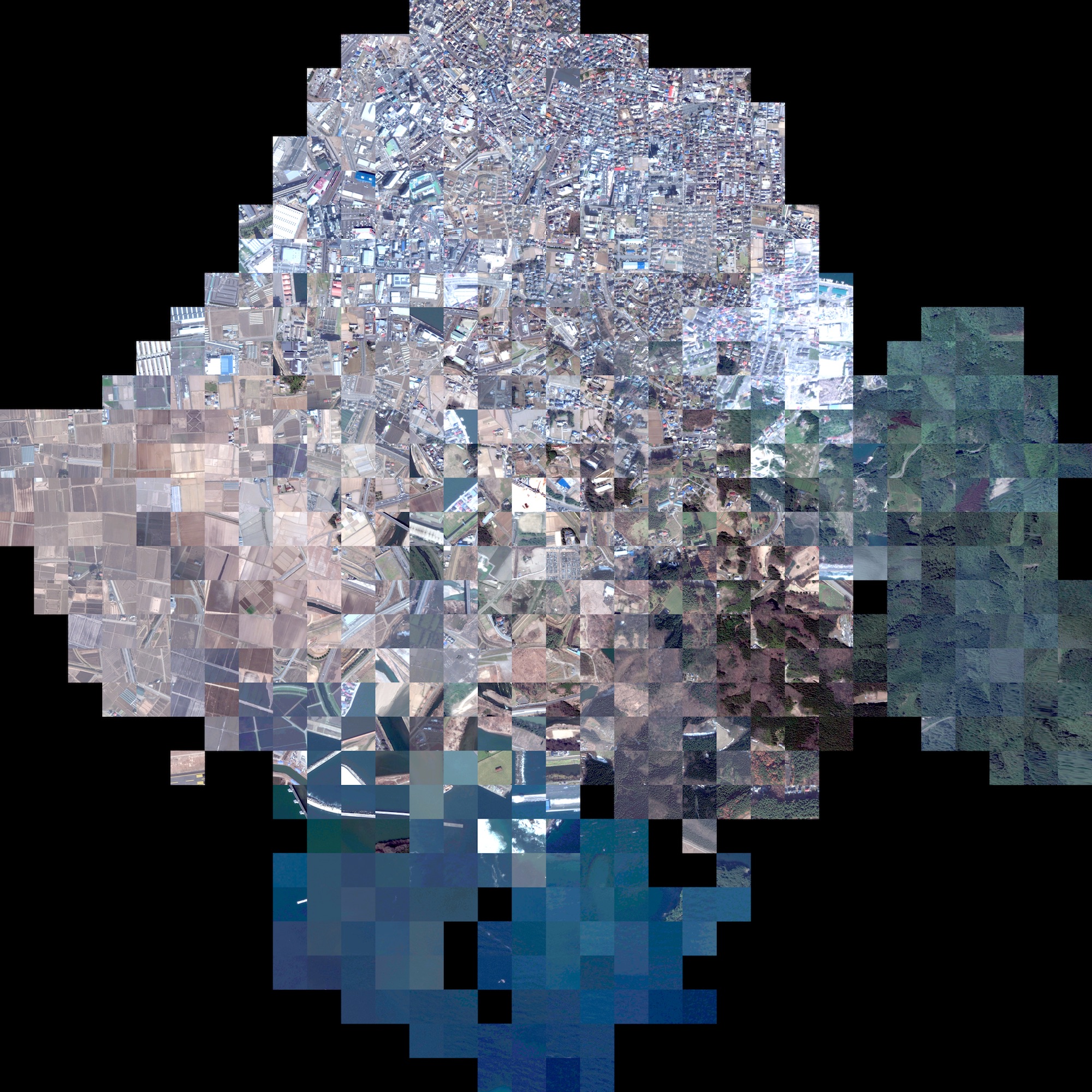}\subcaption{}\label{fig:t-sne:imagenet}
&\hspace*{0mm}\includegraphics[width=80mm,bb=0 0 2000 2000]{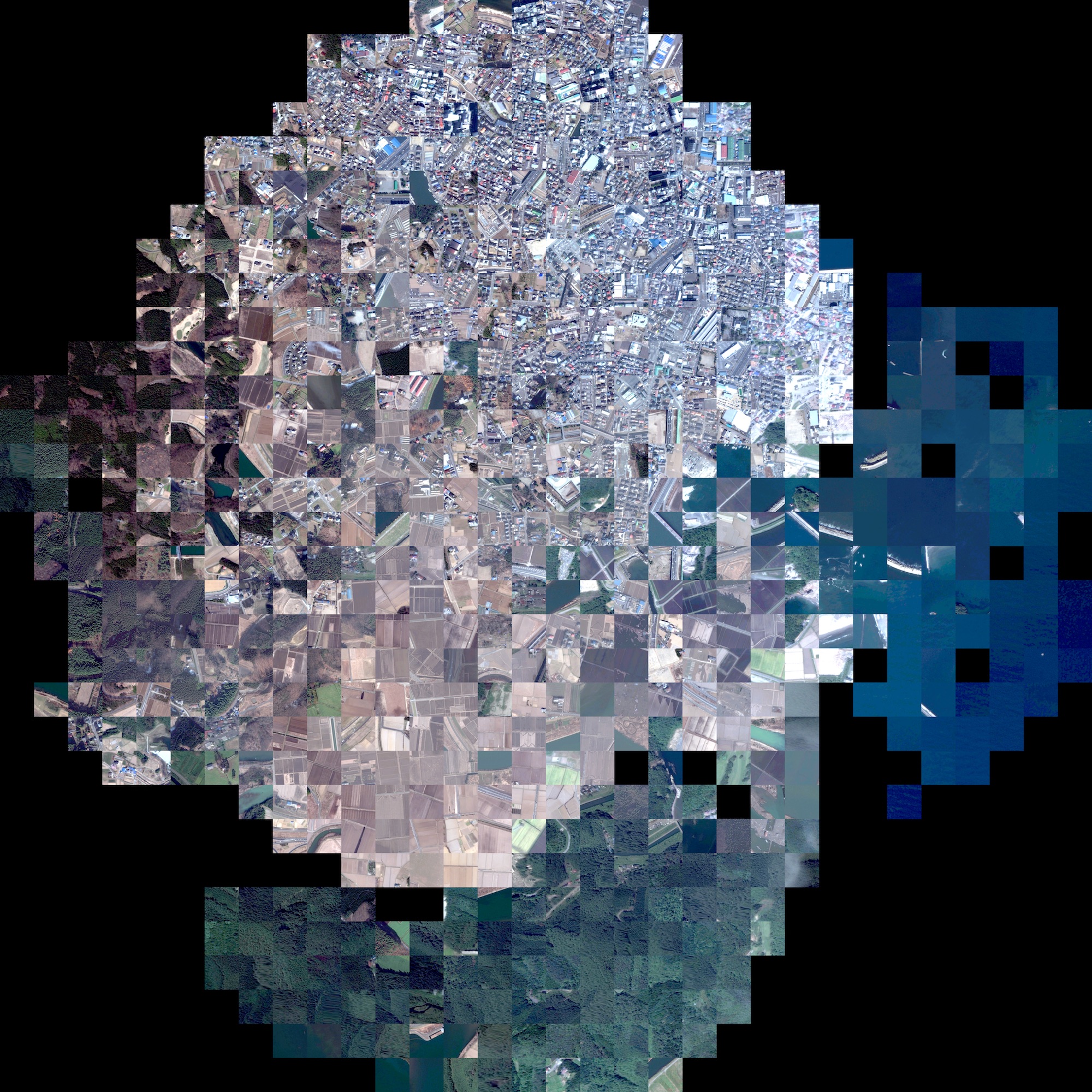}\subcaption{}\label{fig:t-sne:uc}
\end{array}$
\vspace{-3mm}
\caption{Visualization by t-SNE. \protect\subref{fig:t-sne:imagenet}:ImageNet \protect\cite{deng2009imagenet}. Images of urban areas are clustered in the upper region, forests images are clustered at the right, images of the sea are clustered in the lower region, ant the farmlands are clustered  at the left. \protect\subref{fig:t-sne:uc}:UC Merced Land Use Dataset \protect\cite{yang2010bag}. Some images from the same category are distributed separately, for example images of forests are divided into the left and  the lower parts. }
\label{fig:t-sne}

\end{center}
\end{figure*}

\section{Dataset Generation for Cloud Removal}
\label{sec:cloud_dataset}
In this work, images captured by the WorldView-2 earth observation satellite are used. Both visible light images and the NIR images possess the resolution of $20,000 \times 20,000$ with the spatial resolution of 0.5 [m/pixel].  We chose eight comparatively cloudless images, which mainly captured urban areas, for actual learning.  In total, 37,000 images with a resolution of $256 \times 256$ are extracted for training McGANs.

\subsection{Synthesis of cloud-obscured images}
Both cloud obscured images and cloud-free images are indispensable  to train the networks for cloud removal, as they form the training and  ground truth data respectively. However, the appearance varies greatly as the imaging conditions, such as lighting and status on the ground, changes with time even for the same location. Therefore, we create the dataset for learning by synthesizing the simulated cloud on the cloudless or  cloud-free ground truth images. Furthermore, to compensate for the difference in color tone between the cloud synthesized images and the original images, color correction \cite{hunt2005reproduction,egst20141035} is performed on both images.

In this work, the clouds are simulated by Perlin noise \cite{perlin2002improving} firstly. Then the simulated  clouds are combined with the RGB images by alpha blending to generate obscured images. Fig.\ref{fig:synthesis} shows an example of the image synthesis process. The RGB image (Fig.\ref{fig:synthesis:rgb}) is  overlaid by a Perlin noise simulated cloud (Fig.\ref{fig:synthesis:cloud}) with the alpha blending method to synthesize the image (Fig.\ref{fig:synthesis:syn}). Then generated image is further processed by color correction (Fig. \ref{fig:synthesis:cc}). To show the necessity of color correction, we take another image (Fig.\ref{fig:synthesis_colorcorrect}) for comparison. Fig.\ref{fig:synthesis_colorcorrect:before} is the original RGB image of a different location from that in Fig.\ref{fig:synthesis}. The color corrected result is shown in Fig.\ref{fig:synthesis_colorcorrect:after}. By comparing the two groups of images, we can observe that the variety of color tone is greatly improved with the process of color correction.

\subsection{Uniformization of the dataset with t-SNE}
Since most of the earth is covered with seas and forests, the contents of the satellite images used in the work are also mainly of these two types. If we randomly sample the images for training, the learning result is prone to overfitting in certain categories due to the bias of the training data. Hence, we utilized t-SNE to sample the images by categories to avoid this problem.

First, we extract a feature vector of 4096 dimensions from each image with the AlexNet \cite{NIPS2012_4824}. The extracted feature vectors are mapped to the 2D space with t-SNE. Then, we uniformly sample 2000 images from the 2D feature space to create the training dataset.

The ImageNet Large Scale Visual Recognition Challenge (ILSVRC) dataset \cite{deng2009imagenet} and the land use image dataset UC Merced land use dataset \cite{yang2010bag} (21 classes and 100 images for each class) are used for training the AlexNet. The processed results of the features from the two datasets after using t-SNE are shown in Fig.\ref{fig:t-sne}. Fig.\ref{fig:t-sne:imagenet} shows the distribution of the training images mapped with features from ImageNet dataset, and Fig.\ref{fig:t-sne:uc} shows the result with features from UC Merced land use dataset. In the Fig.\ref{fig:t-sne:imagenet}, images of the urban areas are clustered in the upper region, forest images are clustered at the right, images of the sea are clustered  in the lower region and images of farmlands are clustered  at the left. We can see that the images are well clustered by their categories. We also can see a similar result in Fig. \ref{fig:t-sne:uc} except that some images from the same category are distributed separately, e.g., images of forests are divided into the left and the lower parts. This is probably caused by the differences between the images used in this work and the images in UC dataset, in addition to the insufficiency of the images in the UC dataset. Therefore, we adopt the features extracted by the AlexNet from ImageNet for t-SNE.

\begin{figure}[htpb]
\begin{center}
\hspace*{0mm}\includegraphics[width=83mm,bb=0 0 1041 935]{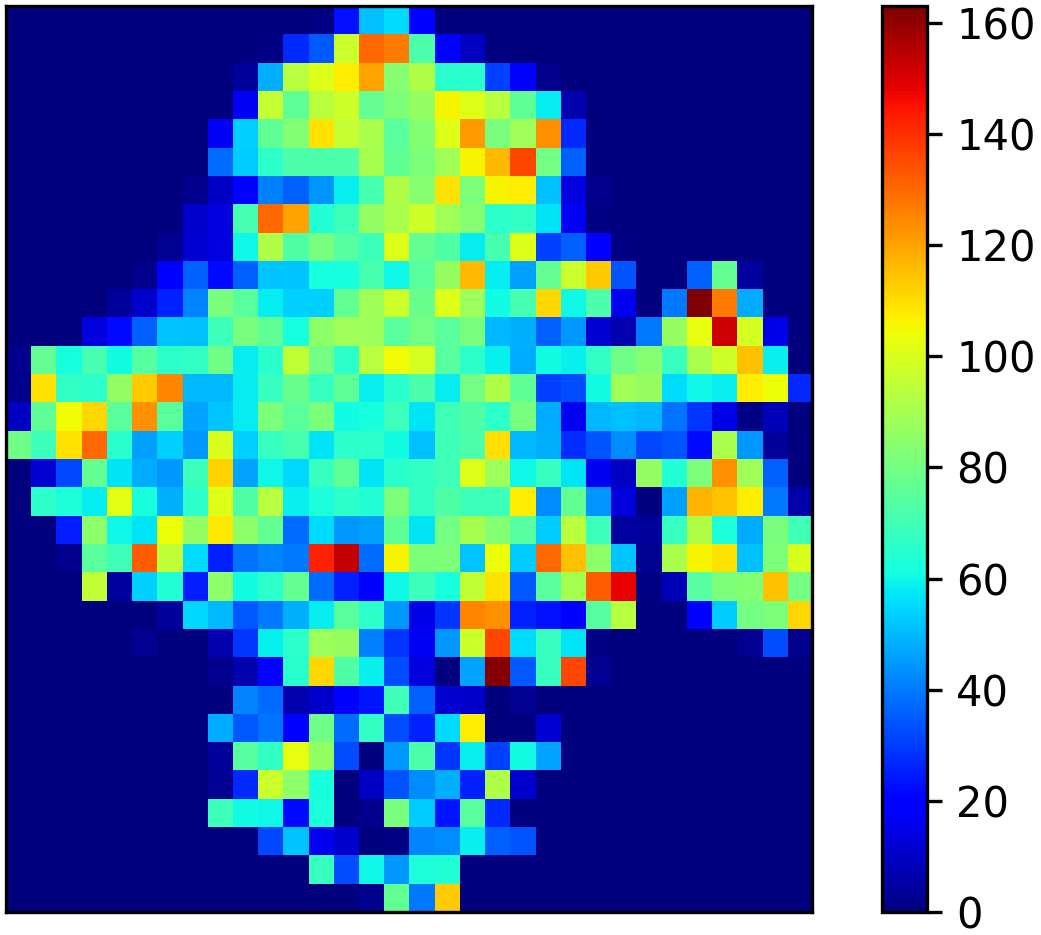}
\caption{Heat map of image distribution mapped by t-SNE. The colors indicate the number of images in the corresponding 2D feature space.}
\label{fig:hist_t-SNE}
\end{center}
\end{figure}

The number of images in each cluster is shown in a heat map in Fig.\ref{fig:hist_t-SNE}. From Fig.\ref{fig:hist_t-SNE} we can see that the images are uniformly distributed except in some regions of the grids. Images are uniformly sampled by the grid to improve the overfitting caused by the bias of in the training data.

\section{Multispectral conditional Generative Adversarial Networks (McGANs)}
\begin{figure*}[htpb]
\begin{center}
\hspace*{0mm}\includegraphics[width=175mm,bb=0 0 1213 381]{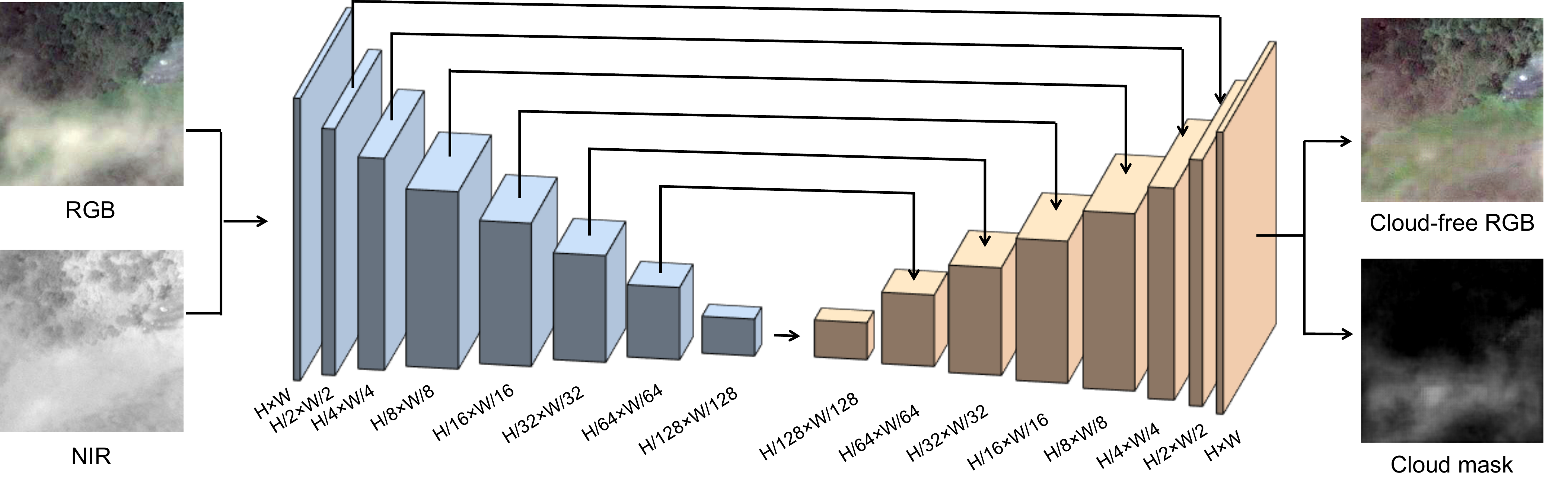}
\caption{Network Architecture of Generator}
\label{fig:generator}
\end{center}
\end{figure*}

In this paper, we propose Multispectral conditional Generative Adversarial Networks (McGANs), which extends the input of cGANs to multispectral images in order to be capable of merging input visible light images and images of longer wavelengths to remove clouds from the visible light images. The detailed architecture of McGANs are shown in Fig. \ref{fig:generator} and Tab.\ref{tab:CNN_Structure}.

We extend the input of the cGANs model proposed in \cite{isola2016image} to four channels RGB-NIR images \footnote{By adding images captured using other wavelength, such as far infrared rays and microwaves, the input will be further extended}. Furthermore, the output is also extended to a total number of four channels, including the predicted RGB image after cloud removal and the grayscale mask image, which is estimated simultaneously to improve the prediction accuracy.
The input RGB-NIR image, the output RGB image and the cloud mask image are normalized to $[-1, 1]$ at each channel and then transferred to the network.

\begin{table}[tb]
\begin{center}
\caption{Network Architecture of McGANs\label{tab:CNN_Structure}}{
\scalebox{1.0}[1.0]{
\begin{tabular}{c|c|c}
\hline
Encoder & Decoder & Discriminator\\ \hline
CR~(64, 3, 1)  & CBRD~(512, 4, 2) & CBR~(64, 4, 2) \\
CBR~(128, 4, 2) & CBRD~(512, 4, 2) & CBR~(128, 4, 2)\\ 
CBR~(256, 4, 2) & CBRD~(512, 4, 2) & CBR~(256, 4, 2)\\ 
CBR~(512, 4, 2) & CBR~(512, 4, 2) & CBR~(512, 4, 2)\\
CBR~(512, 4, 2) & CBR~(256, 4, 2) & C~(1, 3, 1)\\
CBR~(512, 4, 2) & CBR~(128, 4, 2) & \\
CBR~(512, 4, 2) & CBR~(64, 4, 2) & \\
CBR~(512, 4, 2) & C~(4, 3, 1) & \\ \hline
\end{tabular} 
}
}
\end{center}
\end{table}

\subsection*{Network Architecture}
Details of the network structure about McGANs used in this work are shown in Tab.\ref{tab:CNN_Structure}.
The layer of Convolution, Batch Normalization, and ReLU are represented by C, B, R respectively. D indicates that the Dropout is applied.
Numbers in parentheses indicate the number, size, stride of the convolution filters sequentially. In addition, Leaky ReLU is used in all ReLU layers of Encoder and Discriminator.

The objective of a conditional GAN can be expressed as

\begin{equation}
\begin{split}
\mathcal{L}_{cGAN}(G,D)=&\mathbb{E}_{x,y\sim p_{data}(x,y)}[\log D(x,y)]+\\
&\mathbb{E}_{x\sim p_{data}(x),z\sim p_{z}(z)}[\log (1-D(x,G(x,z)))],
\end{split}
\label{eq:gan_loss}
\end{equation}
where Generator $G$ tries to minimize the objective against an adversarial Discriminator $D$ that tries to maximize it.
To encourage less blurring, L1 loss can be added to the objective as follows \cite{isola2016image}
\begin{equation}
G^* = \arg \min_{G} \max_{D} \mathcal{L}_{cGAN}(G,D) + \mathcal{L}_{L1}(G).
\label{eq:minmax}
\end{equation}

Let $I_M$ be the input multispectral image and $I_T$ be the target RGB image with a total of four channels, including RGB and the grayscale mask image of the cloud. The L1 Loss function (denoted as $\mathcal{L}_{L1}$) of the Generator is defined in Eq.\ref{eq:l1_loss}. 
$\lambda_c$ represents the weight of each channel for the loss calculation \footnote{$\lambda_c$ is set to $1$ in this work.}, and $\phi(I_M)$ represents the predicted result from the input image $I_M$ from the trained networks.
\begin{equation}
\mathcal{L}_{L1}(G)=\frac{1}{4HW}\sum_{c=1}^{4}\sum_{v=1}^{H}\sum_{u=1}^{W} \lambda_c|I_T^{(u,v,c)}-\phi(I_M)^{(u,v,c)}|_1
\label{eq:l1_loss}
\end{equation}

\begin{figure*}[!t]
\begin{center}
\hspace*{0mm}\includegraphics[width=175mm,bb=0 0 722 453]{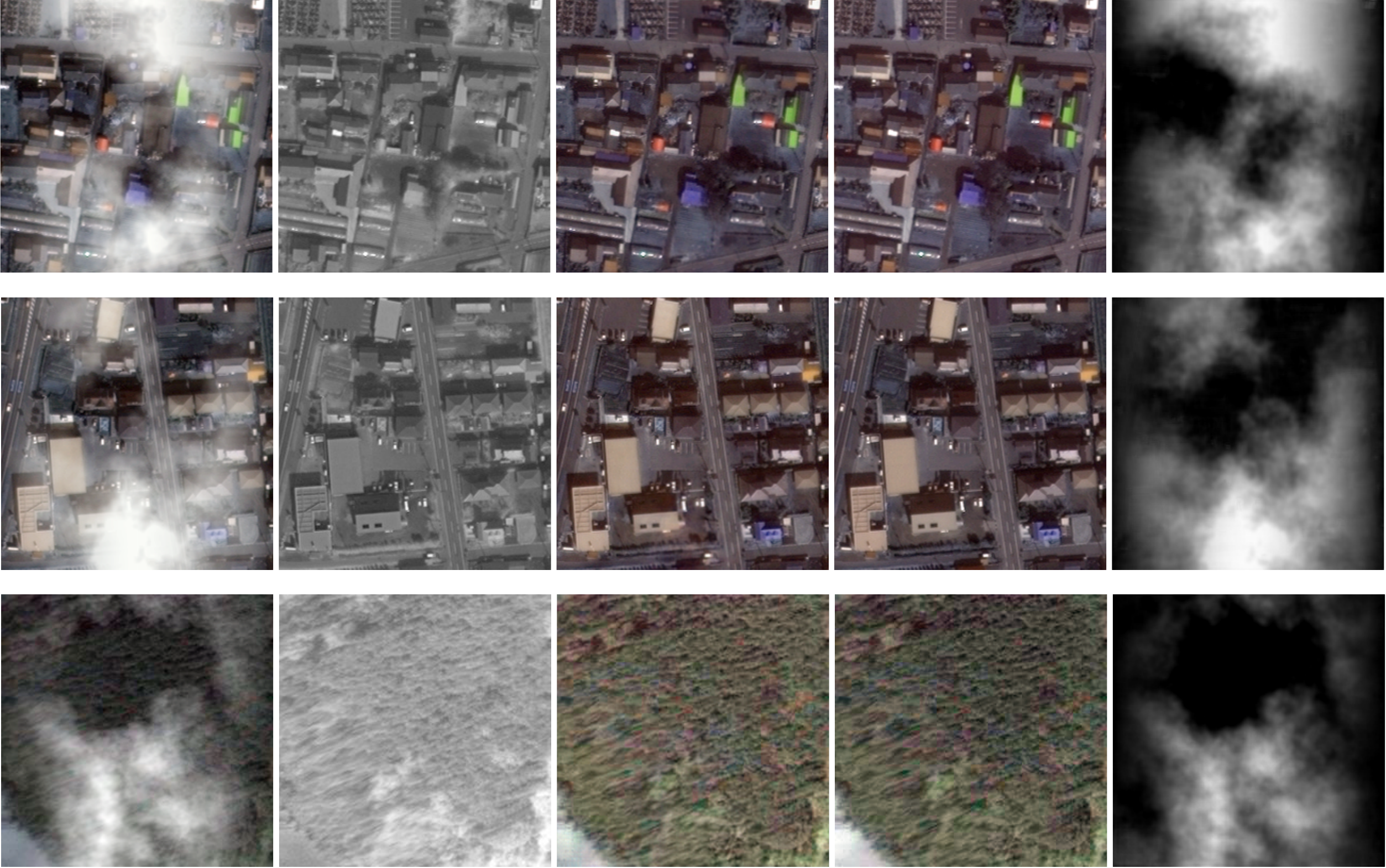}

$\begin{array}{p{32mm}p{32mm}p{32mm}p{32mm}p{32mm}}
 \hspace*{12mm}\raisebox{1mm}{RGB}
&\hspace*{12mm}\raisebox{1mm}{NIR}
&\hspace*{4mm}\raisebox{1mm}{Cloud-free RGB}
&\hspace*{5mm}\raisebox{1mm}{Ground truth}
&\hspace*{5mm}\raisebox{1mm}{Cloud mask}
\end{array}$
\caption{Prediction results by McGANs with the synthesized cloud images}
\label{fig:result02}
\end{center}
\end{figure*}

\begin{figure*}[htb]
\begin{center}
\vspace{-5mm}
\hspace*{0mm}\includegraphics[width=175mm,bb=0 0 722 453]{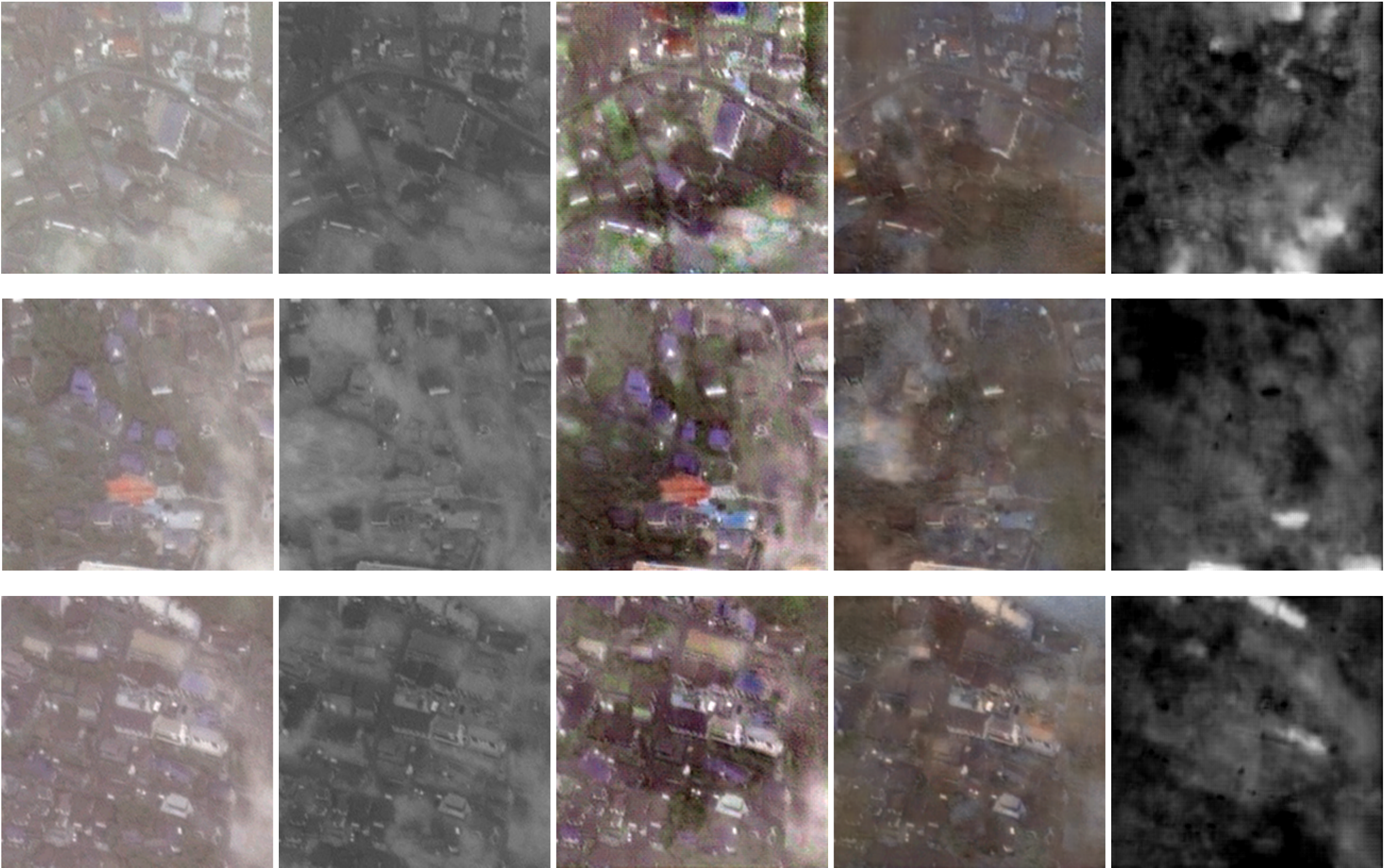}
\vspace{-2mm}
$\begin{array}{p{32mm}p{32mm}p{32mm}p{32mm}p{32mm}}
 \hspace*{12mm}\raisebox{1mm}{RGB}
&\hspace*{12mm}\raisebox{1mm}{NIR}
&\hspace*{4mm}\raisebox{1mm}{Cloud-free RGB}
&\hspace*{8mm}\raisebox{1mm}{NIR2RGB}
&\hspace*{5mm}\raisebox{1mm}{Cloud mask}
\end{array}$
\caption{Prediction results by McGANs with real cloud images}
\label{fig:result01}
\end{center}
\end{figure*}

\section{Evaluation Results}
To evaluate our proposed method, experimental results are listed and discussed in this Section. From the experimental results, we expect to show that the proposed McGANs are able to improve visibility by cloud removal with RGB and NIR satellite images.

As explained earlier, the satellite images captured at different times (even though they might be of the same area), vary greatly in their appearance as imaging conditions, such as lighting and the situation on the ground, change. This makes it difficult to acquire the ground truth of the area blocked by the cloud. We use 5,000 groups of images as described in Sec.\ref{sec:cloud_dataset} to train the network. Each group includes an image of the area not obscured by the cloud, a mask image of the simulated clouds using Perlin noise, a synthesized image and an NIR image. All images are 
 processed with color correction. The number of minibatch is set to 16 and the number of epochs is 500.

To verify the advantage of using multispectral images for cloud removal, we also compare them against the RGB images generated by the networks (NIR-cGANs) with only NIR images as input. NIR images are used as input and images that are not obscured by the cloud are used as ground truth. The same dataset (as in McGANs) is used for training  NIR-cGANs. The number of minibatch and epochs is also the same.

Sample results of the synthesized cloud obscured images are shown in Fig.\ref{fig:result02}. The columns represent the synthesized cloud obscured RGB images, NIR images, RGB images predicted by McGANs, the ground truth and the mask images of the clouds predicted by McGANs, from left to right.

\begin{figure*}[thb]
\begin{center}
\hspace*{0mm}\includegraphics[width=175mm,bb=0 0 721 143]{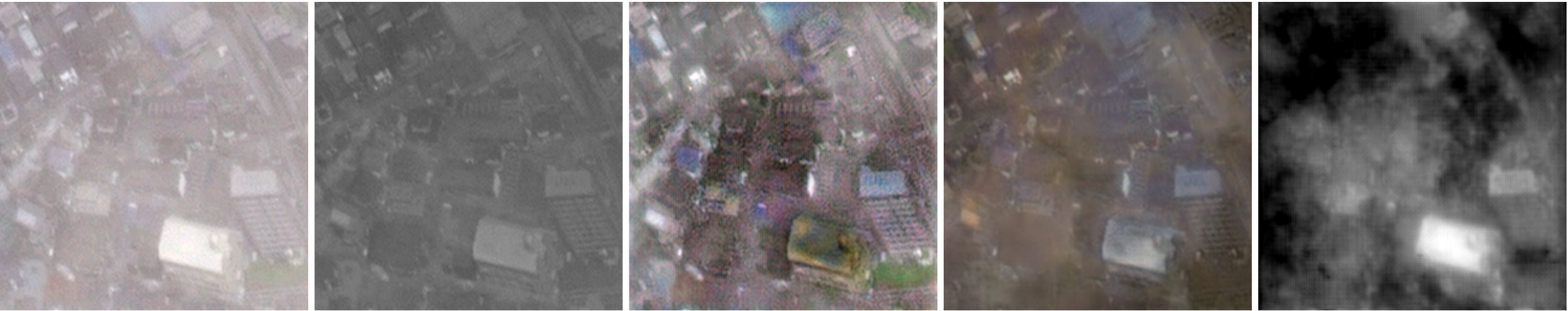}
\vspace{-2mm}
$\begin{array}{p{32mm}p{32mm}p{32mm}p{32mm}p{32mm}}
 \hspace*{12mm}\raisebox{1mm}{RGB}
&\hspace*{12mm}\raisebox{1mm}{NIR}
&\hspace*{4mm}\raisebox{1mm}{Cloud-free RGB}
&\hspace*{8mm}\raisebox{1mm}{NIR2RGB}
&\hspace*{5mm}\raisebox{1mm}{Cloud mask}
\end{array}$
\caption{Failure case due to a white object}
\label{fig:fail01}
\end{center}
\end{figure*}

\begin{figure*}[h!]
\begin{center}
\hspace*{0mm}\includegraphics[width=175mm,bb=0 0 721 143]{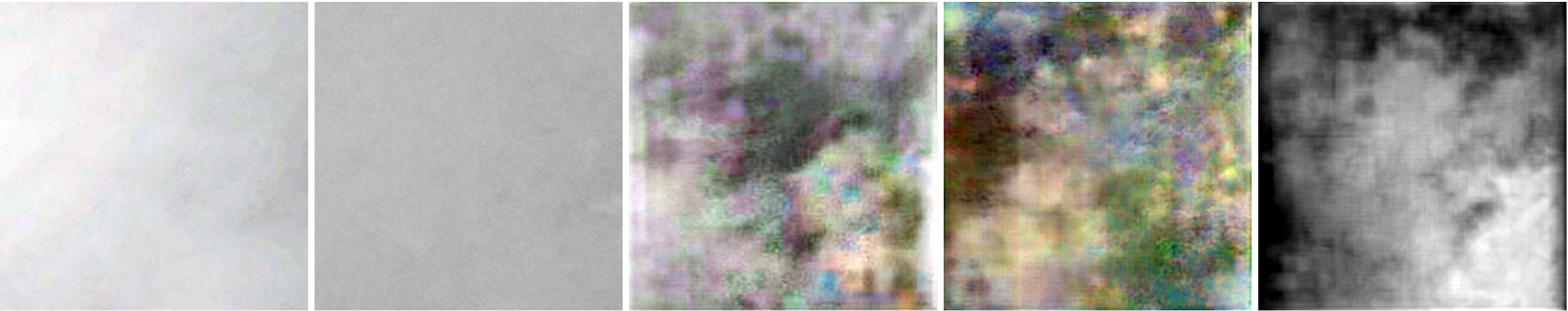}
\vspace{-2mm}
$\begin{array}{p{32mm}p{32mm}p{32mm}p{32mm}p{32mm}}
 \hspace*{12mm}\raisebox{1mm}{RGB}
&\hspace*{12mm}\raisebox{1mm}{NIR}
&\hspace*{4mm}\raisebox{1mm}{Cloud-free RGB}
&\hspace*{8mm}\raisebox{1mm}{NIR2RGB}
&\hspace*{5mm}\raisebox{1mm}{Cloud mask}
\end{array}$
\caption{Thick cloud case}
\label{fig:fail02}
\vspace{-3mm}
\end{center}
\end{figure*}

Sample results of real cloud obscured images are shown in Fig.\ref{fig:result01}, Fig.\ref{fig:fail01} and Fig.\ref{fig:fail02}. The columns represents RGB images obscured by the cloud, NIR images, RGB images predicted by McGANs, RGB images predicted by NIR-GANs and the mask images of the clouds predicted by McGANs, from left to right.
From Fig.\ref{fig:result01}, we can observe that although the images, which are generated  only with NIR images, look like visible light RGB images, their colors differ from the ground truth.
While the clouds are well removed in the predicted results by McGANs except for the region obscured by the cloud that infrared can not penetrate. Even for these regions in the predicted images, the color appears similarly to the very light color in the input images. This also proves that McGANs dose not predict color from the information only  from the NIR images.

\begin{figure*}[h!]
\begin{center}
\hspace*{0mm}\includegraphics[width=175mm,bb=0 0 721 143]{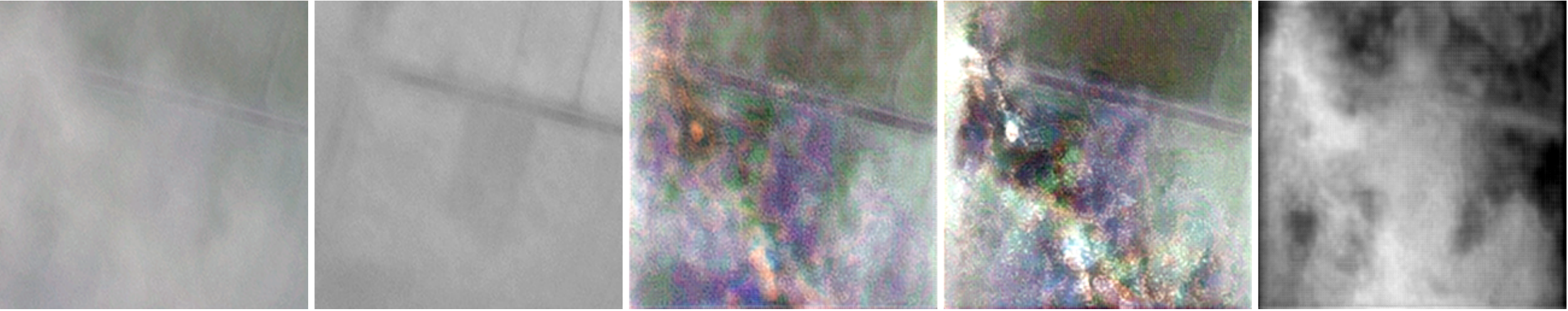}
\vspace{-2mm}
$\begin{array}{p{32mm}p{32mm}p{32mm}p{32mm}p{32mm}}
 \hspace*{12mm}\raisebox{1mm}{RGB}
&\hspace*{12mm}\raisebox{1mm}{NIR}
&\hspace*{4mm}\raisebox{1mm}{Cloud-free RGB}
&\hspace*{8mm}\raisebox{1mm}{RGB2RGB}
&\hspace*{5mm}\raisebox{1mm}{Cloud mask}
\end{array}$
\caption{A prediction result generated from only a RGB image}
\label{fig:rgb2rgb}
\vspace{-3mm}
\end{center}
\end{figure*}

On the other hand, in Fig.\ref{fig:fail01}, we can see that the white object is erroneously recognized as the cloud from the output mask image. This indicates that it is difficult to separate the cloud from the white object with only the visible light images and NIR images, when they are overlapped.
In addition, as seen in Fig.\ref{fig:fail02}, clouds are not removed when they are too thick to be penetrated by NIR. The purpose of this research is to observe the real situation on the ground. Thus, the regions blocked by clouds in the NIR image will not be predicted, which is different from \cite {pathak2016context}. 
When predicting the area blocked by clouds in both visible light and NIR images, it is necessary to model the cloud penetration of NIR based on the visible light images, process the simulated cloud with the penetration model and then synthesize the modeled cloud on the NIR images.
To verify the necessity of the NIR images, we also compare the results generated by our proposed method with that generated from only a RGB image as the input. For thin clouds that can be partly penetrated by the visible light, the results dose not differ much. However, for clouds that can be only penetrated by NIR light, the result with the presence of NIR appears more natural as shown in Fig.\ref{fig:rgb2rgb}. We can see some line contours of the roads on the ground from the upper left part of the NIR image in Fig.\ref{fig:rgb2rgb}, while these contours are occluded in the RGB image. This can be considered as the reason why the result generated with both the NIR and RGB images looks more natural that generated with the RGB image.

From the above results, we have confirmed that the proposed McGANs can remove clouds and predict the color properly when the cloud is thin enough to be penetrated by the NIR.

\section{Conclusion}
In this paper, we have proposed a method to remove thin clouds from satellite images formed using visible light by extending cGANs to multispectral images. 
The dataset for training networks is constructed by synthesizing simulated clouds with Perlin noise over images without clouds, which makes it possible to generate cloud obscured training images and ground truth of the same area.  
In addition, to avoid overfitting to certain categories caused by biased datasets, we introduce t-SNE to sample images uniformly in each category. 
Finally, the experimental results evaluated on the constructed data prove that the clouds in the visible light images can be removed if they are penetrated in NIR images.

In the future, we will extend McGANs to far infrared (FIR) images and SAR images which captured by longer wavelengths and build the networks that can remove all the clouds in the visible light images. 
The findings obtained by analyzing the filters of McGANs in this work can also be applied to establish the model of cloud penetration for waves at each wavelength region or to the physical model of SAR. 
In addition, the simulated clouds with Perlin noise used in this work are somewhat different from real clouds in visible light images. 
Therefore, statistical analysis of actual cloud images is necessary to improve the reality of the simulated clouds for training data. 
Furthermore, we aim to improve the prediction accuracy for different areas by increasing the number and variety of images.


\bibliographystyle{ieee}
\bibliography{egbib}

\end{document}